%% file: 00_main.tex
\documentclass[10pt,twocolumn,letterpaper]{article}

\usepackage{cvpr}              

\usepackage{graphicx}
\usepackage{amsmath}
\usepackage{amssymb}
\usepackage{booktabs}
\usepackage[table,xcdraw]{xcolor}         
\usepackage{multirow}

\usepackage[pagebackref,breaklinks,colorlinks]{hyperref}
\usepackage[accsupp]{axessibility}  

\usepackage[capitalize]{cleveref}
\crefname{section}{Sec.}{Secs.}
\Crefname{section}{Section}{Sections}
\Crefname{table}{Table}{Tables}
\crefname{table}{Tab.}{Tabs.}

\definecolor{Gray}{gray}{0.9}

\DeclareMathOperator*{\mini}{minimize}

\def\cvprPaperID{4860} 
\def\confName{CVPR}
\def\confYear{2023}

\begin{document}

\title{Compositor: Bottom-up Clustering and Compositing for Robust Part and Object Segmentation}

\author{
	Ju He$^1$\footnotemark[1]
	\;\; Jieneng Chen$^1$\footnotemark[1]
	\;\; Ming-Xian Lin$^2$
	\;\; Qihang Yu$^1$
	\;\; Alan Yuille$^1$ \\
	$^1$Johns Hopkins University \;\; $^2$Chinese Academy of Sciences
}
\maketitle
\renewcommand{\thefootnote}{\fnsymbol{footnote}}
\footnotetext[1]{These authors contributed equally to this work.}

\input{01_abstract}
\input{02_introduction}

\input{03_related}
\input{04_method}

\input{05_experiment}
\input{06_conclusion}

{\small
\bibliographystyle{ieee_fullname}
\bibliography{egbib}
}

\end{document}

%% file: 01_abstract.tex
\begin{abstract}
    In this work, we present a robust approach for joint part and object segmentation.
    Specifically, we reformulate object and part segmentation as an optimization problem and build a hierarchical feature representation including pixel, part, and object-level embeddings to solve it in a bottom-up clustering manner. Pixels are grouped into several clusters where the part-level embeddings serve as cluster centers. Afterwards, object masks are obtained by compositing the part proposals.
    This bottom-up interaction is shown to be effective in integrating information from lower semantic levels to higher semantic levels. Based on that, our novel approach \textbf{Compositor} produces part and object segmentation masks simultaneously while improving the mask quality. Compositor achieves state-of-the-art performance on PartImageNet and Pascal-Part by outperforming previous methods by around $0.9\%$ and $1.3\%$ on PartImageNet, $0.4\%$ and $1.7\%$ on Pascal-Part in terms of part and object mIoU and demonstrates better robustness against occlusion by around $4.4\%$ and $7.1\%$ on part and object respectively. Code at \href{https://github.com/TACJu/Compositor}{https://github.com/TACJu/Compositor}.
\end{abstract}

%% file: 02_introduction.tex
\section{Introduction}
Detecting objects and parsing them into semantic parts is a fundamental ability of human visual system. When viewing images, humans not only detect, segment, and classify objects but also segment their semantic parts and identify them. This gives a hierarchical representation that enables a detailed and interpretable understanding of the object which is useful for downstream tasks. For example, humans can estimate the pose of a tiger based on the spatial configuration of its parts and hence judge whether it is about to attack or if it is peacefully sleeping. It is conjectured by cognitive psychologists \cite{biederman1987recognition, lake2015human} that these hierarchical representations are constructed in a \textit{bottom-up} manner where humans first perceive parts and then group them together to form objects. 




By contrast, the computer vision literature on semantic segmentation mostly concentrates on object-level, neglecting intermediate part representations, although object and part segmentation have been shown to be mutually beneficial to each other \cite{eslami2012generative, wang2015joint}. We emphasize that parts help many other tasks such as pose estimation \cite{yang2011articulated, dong2014towards}, detection \cite{azizpour2012object, chen2014detect}, fine-grained recognition \cite{zhang2014part} and few-shot learning \cite{he2023corl}. In addition, exploiting part information can increase robustness of object models against occlusion \cite{kortylewski2020compositional, wang2020robust, wang2020NeMo}. 
Recently, He et al. \cite{he2022partimagenet} proposed PartImageNet, where both part and object annotations are provided. 
Meanwhile, their studies showed that naively using part annotation as deep supervision can improve object segmentation. This motivates us to further design a better interaction pipeline between objects and parts for high-quality segmentation. 



\begin{figure*}[t!]
    \centering
    \includegraphics[width=\textwidth]{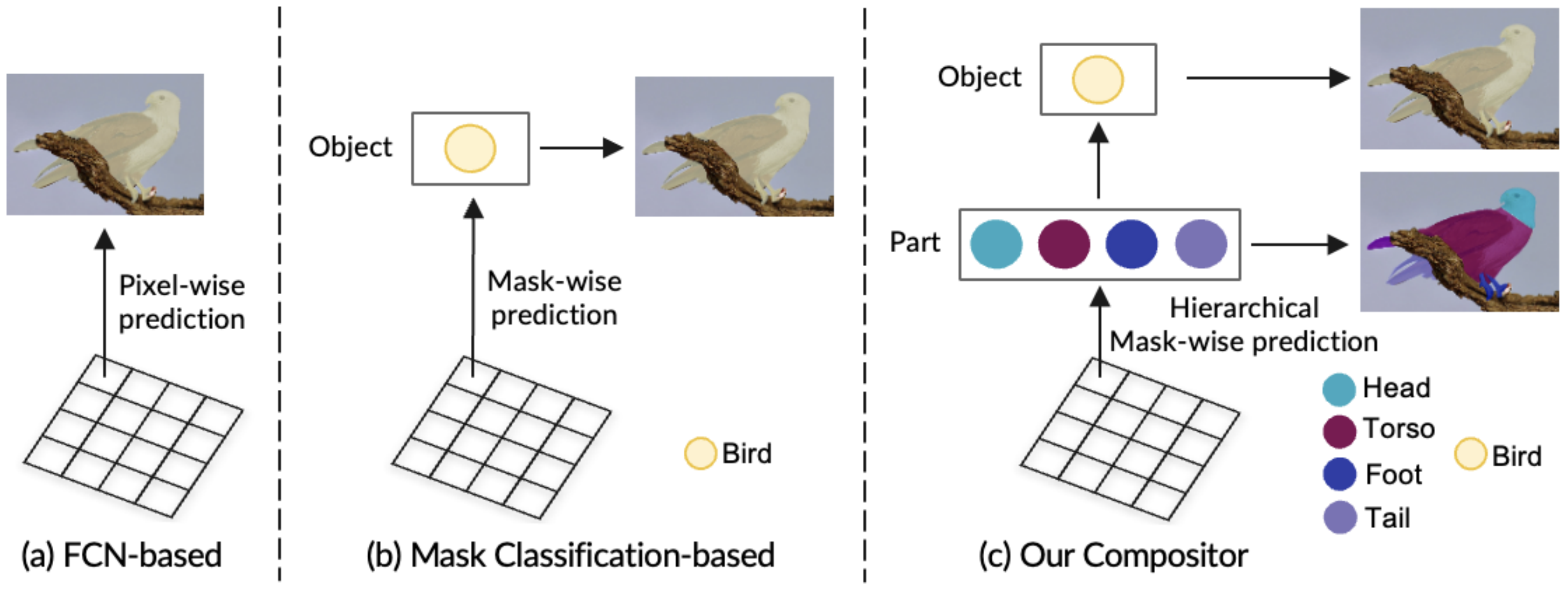}
    \caption{Paradigm comparison among traditional FCN-based method, Mask Classification-based method, and our proposed Compositor for object segmentation. We show example with single object instance here for simplicity.}
    \label{fig:tiser}
\end{figure*}

In this work, we present a strategy for jointly segmenting parts and objects in a bottom-up process. Specifically, we consider a hierarchical representation of images in terms of pixels, parts, and objects. We learn feature embeddings which enables us to reformulate semantic segmentation as an optimization problem whose goal is to find feature centroids that represent parts and objects. As shown in Figure \ref{fig:tiser}, our method uses a bottom-up strategy where pixels are grouped to form part embeddings which, in turn, are grouped to form object embeddings. We implement this in two steps. First, we cluster image pixels to make proposals for object parts. Here the feature embeddings are learned so that pixels belonging to the same part have similar features. Second, we use a similar approach to compose these part proposals to segment the whole object which involves selecting some part proposals and rejecting others. Our complete algorithm, \textit{Compositor}, for segmenting parts and objects consists of these \textit{clustering} and \textit{compositing} steps. This novel algorithm not only helps us to build a hierarchical segmentation model but also increases the robustness of the model against occlusion since our parts are clustered based on the similarity of pixel features, which are less affected by occlusion compared to other context-based methods. Moreover, objects are constructed using parts that helps minimize the influence of occlusion.

We verify Compositor's effectiveness on both PartImageNet \cite{he2022partimagenet} and Pascal-Part \cite{chen2014detect}, where the former focuses on single-instance and the latter contains more multi-instances scenarios. We show that Compositor generates high-quality semantic parts from pixels which further benefits object segmentation. Quantitatively, Compositor achieves $61.44\%$ and $71.78\%$ mIoU on part and object segmentation with the ResNet-50 \cite{he2016deep}, outperforming single-task specialized MaskFormer \cite{cheng2021per} by $1.1\%$ and $1.6\%$ respectively. We get consistent improvement on Pascal-Part by surpassing MaskFormer by $0.4\%$ and $1.7\%$ in terms of part and object mIoU. 

We further show the robustness of Compositor against occlusion with Occluded-PartImageNet, which is obtained by appending artificial occluders on the original images in PartImageNet following the protocol of OccludedPASCAL3D+ \cite{wang2020robust}. As a result, Compositor outperforms MaskFormer by around $4.4\%$ and $7.1\%$ on part and object mIoU respectively. Ablation studies are conducted to validate the effectiveness of our key designs. Qualitative visualization results on both clean images and occluded images are presented. Error analysis is conducted to better understand the model and guide future work. In summary, we make the following contributions in this work:


\begin{enumerate}
    \item We propose a bottom-up strategy for segmentation, where we first generate parts from pixels followed by compositing parts into objects. This strategy gives us a joint solution for part and object segmentation.
    \item We validate Compositor on PartImageNet and Pascal-Part by extensive experiments showing that interactions between parts and objects help each other and result in state-of-the-art performance on both tasks.
    \item We create Occluded-PartImageNet by adding occluders enabling us to demonstrate the innate robustness of Compositor against occlusion. 
\end{enumerate}

%% file: 03_related.tex
\section{Related Works}

\subsection{Object Parsing}
Parsing objects into parts is a long-standing problem in computer vision and there is a rich literature on the topic. Pictorial Structure was first proposed in the early 1970's \cite{fischler1973representation}. After that, plenty of different methods \cite{weber2000unsupervised, felzenszwalb2005pictorial, fei2006one, zhu2007stochastic, girshick2011object} have been proposed to explicitly model parts and their spatial relations to the whole object. These methods share a common theme that the object-part models provide rich representations of objects and help interpretability. In the era of deep learning with data-driven models, research on part-based models gets hindered due to the lack of large-scale datasets. 
Huang et al. \cite{hung2019scops} proposed a self-supervised co-segmentation method for generating semantically consistent part segmentation results on certain objects. Liu et al. \cite{liu2021unsupervised} disentangled object appearance and shape information to learn a part segmentation model in an unsupervised manner. 
However, these works mainly focus on unsupervised part discovery in specific classes instead of finding parts and evaluating them precisely in more general classes.

\subsection{Semantic Segmentation}
Semantic segmentation has been extensively studied and evaluated on multiple benchmarks. Classic works in semantic segmentation \cite{he2017mask, chen2017deeplab, chen2018encoder} adopt per-pixel classification setting. 
With the recent progress in transformers \cite{carion2020end}, a new paradigm named mask classification~\cite{wang2021max,cheng2021per} has been proposed, where segmentation predictions are represented by a set of binary masks with its class label, which is generated through the conversion of object queries to mask embedding vectors 
followed by multiplying with the image features. The predicted masks are trained by Hungarian matching with ground truth masks. Thus the essential component of mask transformers is the decoder which takes object queries as input and gradually transfers them into mask embedding vectors. Most of the recent works \cite{strudel2021segmenter, wang2020axial, cheng2021per, wang2021max, zhang2021k, cheng2022masked} adopt this setting and the major difference lies in the design of the decoder, while in this work, we propose a novel decoder with \textit{clustering} and \textit{compositing} steps, inspired by the transformer decoder~\cite{vaswani2017attention} and its variants~\cite{li2019expectation,locatello2020object,cheng2022masked,yu2022cmt,yu2022k}, which gives a joint solution for part and object segmentation.

\subsection{Hierarchical learning of objects and parts}
Learning objects through the intermediate representation - parts, is a challenging but attractive research topic as it provides a more robust and interpretable understanding of objects. Morabia et al. \cite{morabia2020attention} first proposed to solve part and object detection simultaneously through an attention mechanism. Recently, Ziegler and Asano \cite{ziegler2022self} shows that learning object parts serves as a good pretext task for self-supervised semantic segmentation as it can provide spatially diverse representation. 
As far as we've concerned, the most related work to ours is Wang et al. \cite{wang2015joint} on the PASCAL-Part \cite{chen2014detect}, which predicted parts and objects simultaneously.
However, in most of these works, object representation still comes through the pixel features, while we show that object representation can be updated based only on part proposals.

\subsection{Robustness against occlusion}
To ensure good performance in real-world conditions it is crucial to evaluate the robustness of vision algorithms in out-of-distribution scenarios. In particular, as proposed by \cite{wang2020robust}, it is important to test robustness to occluders either by creating a new dataset by superimposing objects onto images or by carefully estimating the amount of occlusion that exists in the dataset. Several works focus on improving robustness through architecture changes. For example, analysis-by-synthesis approaches \cite{kortylewski2020compositional, wang2020NeMo, ma2022robust} using compositional networks show much stronger robustness against occlusion based on the generative nature of the model. In this work, we follow the same strategy as \cite{wang2020robust} to create Occluded-PartImageNet and use it to verify the innate robustness of our model in this out-of-distribution setting.

%% file: 04_method.tex
\section{Method}

In this section, we first provide a view that regards the object segmentation problem as an optimization problem of learning a feature vector as a centroid for grouping corresponding pixels. Meanwhile, we introduce a hierarchical representation to tackle the problem from part to whole. Afterwards, we illustrate the proposed Compositor, which clusters pixel embedding to part embedding and further composites part embedding to obtain object embedding. In the end, we discuss how we train the model and how to obtain both part and object predictions during the inference.

\begin{figure*}[t!]
    \centering
    \includegraphics[width=\textwidth]{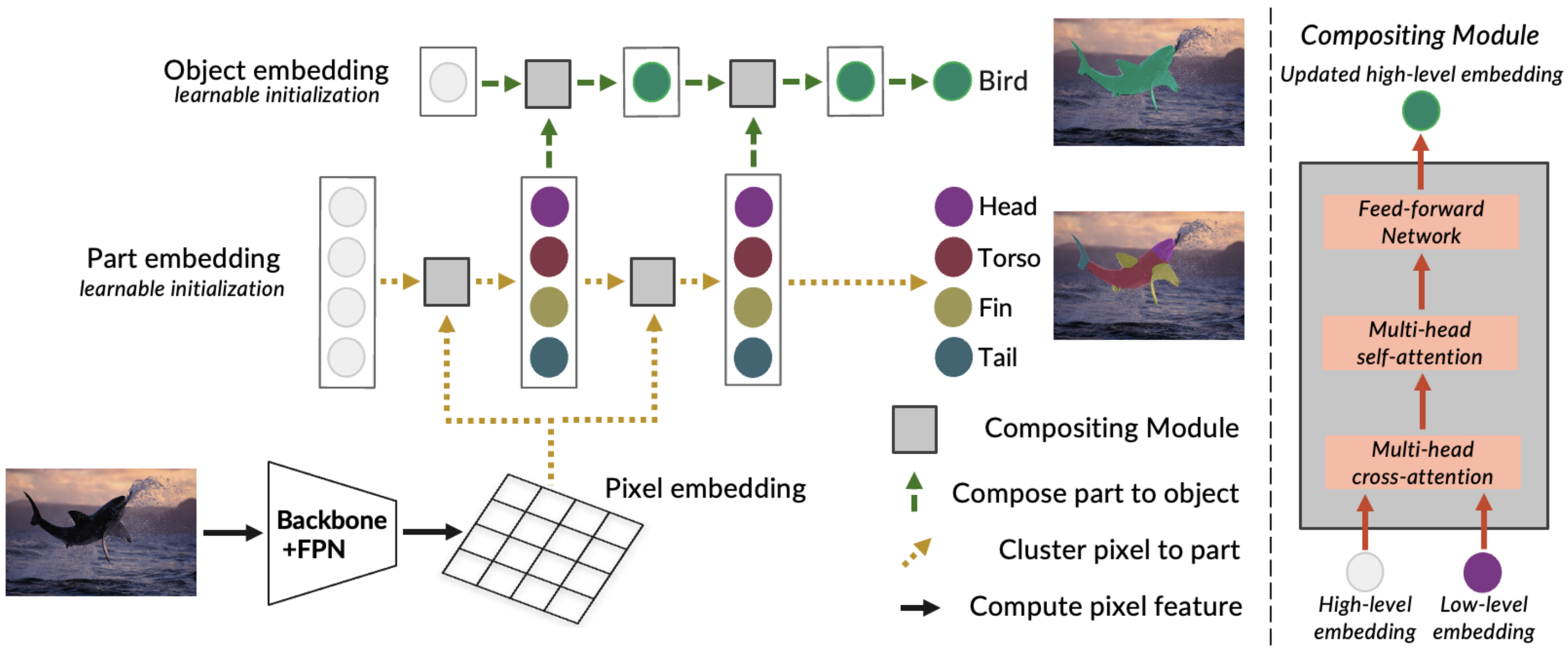}
    \caption{Overview of the proposed framework Compositor, which clusters pixel embedding to part embedding and further composites part embedding to obtain object embedding. Part embedding and object embedding will be iteratively updated and used to predict part/object categories and segmentation maps. 
    }
    \label{fig:framework}
\end{figure*}

\subsection{Segmentation as Clustering and Compositing}

An image $\mathbf{I} \in \mathbb{R}^{H \times W \times 3}$ can be viewed as a set of non-overlapping regions with associated labels:
\begin{equation}
    \{y_i\}_{i=1}^M = \{(d_i, c_i)\}_{i=1}^M \,
    \label{formula:mask_def}
\end{equation}
where $d_i \in {\{0,1\}}^{H \times W}$ indicates whether a pixel belongs to the region and $c_i$ denotes the class label of region $d_i$. $M$ is the number of non-overlapping regions.

Segmentation aims to segment the image into such regions. Two mainstream methods exist to tackle this problem. One directly gives the per-pixel prediction with a prediction head in FCN manner~\cite{long2015fully, chen2017deeplab}. Another one, inspired by the transformers~\cite{vaswani2017attention}, proposes to disentangle the problem into class-agnostic mask segmentation and mask classification~\cite{carion2020end, cheng2021per, wang2021max}. In this paper, we present a more general view of the problem, by considering the task of segmentation is essentially the same as learning a feature embedding, which can serve as a centroid to group the corresponding pixels together and thus formulates a mask. We first define a bottom-up representation, consisting of pixels, parts, and objects as:

\textbf{Pixel embedding}. Given an image $\mathbf{I} \in \mathbb{R}^{H \times W \times 3}$, we use a CNN or Transformer backbone to extract the image features, which are then fed into a FPN~\cite{lin2017feature} to obtain the feature embedding in a higher-resolution. We further add a learnable positional encoding on it and forward it into a MLP layer to obtain the final feature embedding $\mathbf{F} \in \mathbb{R}^{HW\times C}$ for pixels, where $C$ is the feature dimension. Formally, we have $\mathbf{F} = f(\mathbf{I};\theta)$, where $\theta$ is the parameters of the backbone, FPN, and MLP.

\textbf{Part embedding}. Part is a natural intermediate representation for objects~\cite{kortylewski2020compositional}. We consider a feature embedding $\mathbf{P} \in \mathbb{R}^{N\times C}$ for object parts with N specifying the number of part proposals, which groups pixels belonging to the same part region. Thus, each $\mathbf{P}_i$ can be viewed as the clustering centroid of the corresponding part pixel features.

\textbf{Object embedding}. Object offers a higher level abstraction of the scene and is usually the most important notion that we care in the scene. Similar to parts, a feature vector $\mathbf{O} \in \mathbb{R}^{M\times C}$ is used to group pixels into object masks.

With the above notations, we reformulate the energy function to minimize for part and object segmentation as:
\begin{equation}
\begin{aligned}
    \centering
    \mini\limits_{w.r.t \ \textbf{F}, \textbf{P}, \textbf{O}} &\sum_{i=1}^{HW}\sum_{a=1}^{N}W_{ia}\cdot\mathrm{Dis}(\mathbf{F}_i,\mathbf{P}_a) 
    \\ + &\sum_{i=1}^{HW}\sum_{b=1}^{M}W_{ib}\cdot\mathrm{Dis}(\mathbf{F}_i,\mathbf{O}_b),
    \label{formula:opt_goal}
\end{aligned}
\end{equation}
where $W_{ia},W_{ib} \in \{0, 1\}$ denote the ground truth mask annotation in a format whether pixel $i$ belongs to the corresponding part $a$ and the object $b$. $\mathrm{Dis}$ is a distance function for measuring the similarity between pixel feature vector $\mathbf{F}$ and part feature embedding $\mathbf{P}$ and object feature embedding $\mathbf{O}$. Note that $\sum_aW_{ia}=1, \sum_bW_{ib}=1$ as each pixel is assigned to only one part/object. In short, the problem has been transferred to learning good feature representations for pixels, parts, and objects, which can then be used to perform clustering and compositing for the segmentation problem.

\subsection{Compositor: Hierarchical Segmentation from Part to Whole}

In prior arts, objects are usually directly derived from pixels while an interpretable intermediate representation is missing. Though \cite{he2022partimagenet} proposes a task to handle object parsing and object segmentation at the same time, object parsing is only used as a deep supervision to help the object segmentation. Nonetheless, we note that this strategy does not make full use of the great potential and relationship between the two tasks. Therefore, we propose our hierarchical framework Compositor, which groups pixels into semantic parts first, and then composites them into objects.

\textbf{Grouping Pixels into Parts.} The aim is to decompose an image into a subset of regions where each region is assigned a part class label. Formally, the model decomposes $I$ into disjoint regions $\hat{D}$, such that $\bigcup _a \hat{d}_a = I$. 
Each $\hat{d}_a$ is assigned a class label $\hat{c}_a \in \{1,...,K\}$ corresponding to either object parts or background. 

We obtain such decomposition by computing the similarity between pixel features $\mathbf{F}$ and part centroids $\mathbf{P}$, thus acquiring the soft-assignment $\hat{W}$ between pixels and corresponding parts. Formally, we estimate the $\mathbf{P}$ and $\hat{W}$ by the following update rules: 
\begin{equation}
    \begin{aligned}
        \hat{W}_{ia} &= {{\exp (\mathbf{P}_a \times (\mathbf{F}_i)^T)}\over{\sum _n \exp (\mathbf{P}_n \times (\mathbf{F}_i)^T)}} \\ 
        \mathbf{P}_a &= \sum _i \hat{W}_{ia} \cdot \mathbf{F}_i,
    \end{aligned}
\end{equation}
which minimize the the first term $\sum\limits_{i=1}^{HW}\sum\limits_{a=1}^{N}W_{ia}\cdot\mathrm{Dis}(\mathbf{F}_i,\mathbf{P}_a)$ in Eq. \ref{formula:opt_goal} with respect to $W$ and $\mathbf{P}$ separately. We further revise the update rules by computing linear projections of the embedding and utilizing a skip-connection on $\mathbf{P}$ inspired by recent success in Transformers:
\begin{equation}
    \begin{aligned}
        \hat{W}_{ia} &= {{\exp (\mathrm{Linear}(\mathbf{P}_a) \times (\mathrm{Linear}(\mathbf{F}_i))^T)}\over{\sum _n \exp (\mathrm{Linear}(\mathbf{P}_n) \times (\mathrm{Linear}(\mathbf{F}_i))^T)}} \\
        \mathbf{P}_a' &= \mathbf{P}_a + \sum _i \hat{W}_{ia} \cdot \mathrm{Linear}(\mathbf{F}_i),
    \end{aligned}
\end{equation}
where $\mathbf{P}_a'$ is the updated version of part embedding. Note that our part embedding initialization $\mathbf{P}_{init}$ is a set of learnable parameters to encourage the part embedding to have the ability to generate part proposals.

With the above update rules, we exploit the part centroids $\mathbf{P}_a$ to generate the mask regions by applying a threshold on the soft-assignment $\hat{W}_{ia}$ to obtain a hard assignment between pixels and part mask $\hat{d}_a$. We further learn a MLP (multi-layer perceptron) with parameter $\phi$ to produce the class label $\hat{c}_a$ for each part mask $\hat{d}_a$. 

\textbf{Compositing Parts into Objects.} Nonetheless, unlike pixels to parts where each pixel will eventually correspond to some part embeddings, the part embeddings may not all be used to obtain the object embedding (\eg, part centroid $\mathbf{P}_a$ that corresponds to bicycle wheel might not have any mask prediction in an image that only contains animals.). To this end, part to object is more like compositing instead of clustering, where each part may be detected or dropped for the current image. Therefore, we propose to first filter out those undetected parts (based on the previous results of part detection) beforehand to only keep a subset $\mathbf{A}$ of the part centroids $\mathbf{P}$ in order to ease the compositing process. We then estimate the object centroids $\mathbf{O}$ based on $\mathbf{A}$ as:
\begin{equation}
    \begin{aligned}
        \hat{W}_{ab} &= {{\exp (\mathrm{Linear}(\mathbf{O}_b) \times (\mathrm{Linear}(\mathbf{A}_a))^T)}\over{\sum _m \exp (\mathrm{Linear}(\mathbf{O}_m) \times (\mathrm{Linear}(\mathbf{A}_a))^T)}} \\
        \mathbf{O}_b' &= \mathbf{O}_b + \sum _a \hat{W}_{ab} \cdot \mathrm{Linear}(\mathbf{A}_a) \\
        \mathbf{A} &= \{\mathbf{P}_a|\text{ if $\mathbf{P}_a$ is detected}\},
        \label{formula:object_update}
    \end{aligned}
\end{equation}
which minimize $\sum\limits_{a=1}^{N}\sum\limits_{b=1}^{M}W_{ab}\cdot\mathrm{Dis}(\mathbf{P}_a,\mathbf{O}_b)$. As the part embeddings serve as the cluster centroids of the pixels belonging to the parts and the object embeddings are composited from the part ones, this optimization goal essentially achieves a similar effect compared to minimizing the second term in Eq. \ref{formula:opt_goal}. We obtain our pixel-object hard assignment by multiplying $\mathbf{O}$ with the image features $\mathbf{F}$ followed by applying a threshold to obtain a binary mask. A MLP with parameter $\psi$ is learned to produce the object class label $\hat{c}_b$ for each predicted object mask $\hat{d}_b$. Object embedding initialization $\mathbf{O}_{init}$ is a set of learnable parameters as well.

In summary, Compositor resolves the object segmentation problem in a hierarchical manner, by clustering pixel embedding into part embedding and then compositing part embedding into object embedding, thus achieving part segmentation simultaneously. The model outputs $\{\mathbf{P}_a, W_{ia}\}$ and $\{\mathbf{O}_b, W_{ab}\}$, which specify the mask regions $\{\hat{d}_a\}$ and $\{\hat{d}_b\}$ with their mask classification labels $\{\hat{c}_a\}$ and $\{\hat{c}_b\}$.

\textbf{Learning the model}. We aim at optimizing the model parameters $\theta, \phi, \psi$ along with $\textbf{P}_{init}$ and $\textbf{O}_{init}$ jointly. Given the mask predictions $\{\hat{d}, \hat{c}\}$ and the groundtruth mask $\{d, c\}$ of the image, we compute an assignment $V$ between the predictions and groundtruth, which can be formulated as a correspondence problem with energy: 
\begin{equation}
    \centering
    \mini\limits_{w.r.t \ V} \sum_{i,j}V_{ij} \mathcal{L}((d_i,c_i), (\hat{d}_j, \hat{c}_j)),
    \label{formula:matching_energy}
\end{equation}
where $\mathcal{L}$ denotes the loss between the predictions and groundtruth. Note that since we typically have much more mask predictions than the groundtruth masks (\ie, $i > j$), we have an additional $\emptyset$ for mapping those unmatched predictions. Except for this $\emptyset$, we have constraints for Eq. \ref{formula:matching_energy} as $\forall i, \sum\limits_j V_{ij} = 1 \ \& \ \forall j, \sum\limits_i V_{ij} = 1$. Inspired by previous works \cite{carion2020end,wang2021max}, this matching process is implemented through Hungarian Matching \cite{kuhn1955hungarian} for efficiently computing the matched pairs. We define our loss function between the matched pairs as $\mathcal{L} = \lambda_{ce}\mathcal{L}_{ce} + \lambda_{dice}\mathcal{L}_{dice} + \lambda_{cls}\mathcal{L}_{cls}$, where $\mathcal{L}_{ce}$ and $\mathcal{L}_{dice}$ compute the binary cross-entropy loss and dice loss \cite{milletari2016v} between $d$ and $\hat{d}$, $\mathcal{L}_{cls}$ computes the classification loss on the label $c, \hat{c}$, $\lambda_{ce}, \lambda_{dice}, \lambda_{cls}$ sets the loss weight respectively. Our final loss consists of loss on both parts and objects and can thus be formulated as:  
\begin{equation}
\begin{aligned}
    \mathcal{L} &= \lambda_{ce}\mathcal{L}_{ce_p} + \lambda_{dice}\mathcal{L}_{dice_p} + \lambda_{cls}\mathcal{L}_{cls_p} \\ &+ \beta(\lambda_{ce}\mathcal{L}_{ce_o} + \lambda_{dice}\mathcal{L}_{dice_o} + \lambda_{cls}\mathcal{L}_{cls_o}),
\end{aligned}
\end{equation}
where the subscript p and o denote parts and objects respectively and $\beta$ is an additional hyper-parameter for controlling the tendency to whether parts or objects of the model.

\textbf{Post-processing.} Ideally, for several categories of objects, each of their corresponding parts should be connected to at least one other object part unless occlusion (\eg, all animal parts should connect). We exploit this weak object knowledge prior as a post-processing protocol to effectively remove portions of false part predictions in the background. To be specific, during inference time, we first search our predicted segmentation mask to find all connected components. We then go over all the components to check if they are connected to other parts and change the label of those isolated ones into background.

%% file: 05_experiment.tex
\begin{table*}[]
    \small
    \centering
    \caption{PartImageNet \textit{val} set results. mIoU, mACC on parts and objects are reported. $\dag$: Models on parts and objects are trained separately, $\ddag$: Models on parts and objects are trained jointly, $*$: Models on objects are trained with parts as deep supervision.}
    \begin{tabular}{l|c|c|cc|cc}
    \multirow{2}{*}{method} & \multirow{2}{*}{backbone} & \multirow{2}{*}{params} & \multicolumn{2}{c|}{Part} & \multicolumn{2}{c}{Object} \\ \cline{4-7} 
     & & & mIoU & mACC & mIoU & mACC \\ \hline
    $\text{Deeplab v3+}^\dag$~\cite{chen2018encoder} & ResNet-50~\cite{he2016deep} & 84M (42M$\times$2) & 60.57 & 71.07 & 68.38 & 81.00 \\
    $\text{Deeplab v3+}^*$~\cite{he2022partimagenet} & ResNet-50~\cite{he2016deep} & 42M & - & - & 69.82 & 81.96 \\
    $\text{MaskFormer}^\dag$~\cite{cheng2021per} & ResNet-50~\cite{he2016deep} & 90M (45M$\times$2) & 60.34 & 72.75 & 70.21 & 81.99 \\
    $\text{MaskFormer-Dual}^{\ddag}$ & ResNet-50~\cite{he2016deep} & 50M & 58.02 & 70.42 & 70.44 & 81.81 \\
    $\textbf{Compositor}^\ddag$ & ResNet-50~\cite{he2016deep} & 50M & \textbf{61.44} & \textbf{73.41} & \textbf{71.78} & \textbf{83.01} \\ \hline
    $\text{SegFormer}^\dag$~\cite{xie2021segformer} & MiT-B2~\cite{xie2021segformer} & 48M (24M$\times$2) & 61.97 & 73.77 & 74.55 & 85.24 \\
    $\text{MaskFormer}^\dag$~\cite{cheng2021per} & Swin-T~\cite{liu2021swin} & 92M (46M$\times$2) & 63.96 & 77.37 & 77.92 & 87.44 \\
    $\text{MaskFormer-Dual}^{\ddag}$ & Swin-T~\cite{liu2021swin} & 51M & 61.69 & 75.64 & 77.24 & 87.12 \\
    $\textbf{Compositor}^\ddag$ & Swin-T~\cite{liu2021swin} & 51M & \textbf{64.64} & \textbf{78.31} & \textbf{78.98} & \textbf{87.80}
    \end{tabular}
    \label{tab:main}
\end{table*}

\begin{table*}[]
    \small
    \centering
    \caption{Pascal-Part \textit{val} set results. mIoU, mACC on parts and objects are reported. $\dag$: Models on parts and objects are trained separately, $\ddag$: Models on parts and objects are trained jointly.}
    \begin{tabular}{l|c|c|cc|cc}
    \multirow{2}{*}{method} & \multirow{2}{*}{backbone} & \multirow{2}{*}{params} & \multicolumn{2}{c|}{Part} & \multicolumn{2}{c}{Object} \\ \cline{4-7} 
     & & & mIoU & mACC & mIoU & mACC \\ \hline
    $\text{MaskFormer}^\dag$~\cite{cheng2021per} & ResNet-50~\cite{he2016deep} & 86M (43M$\times$2) & 47.61 & 58.59 & 72.69 & 81.89 \\
    $\text{MaskFormer-Dual}^{\ddag}$ & ResNet-50~\cite{he2016deep} & 50M & 46.60 & 57.96 & 72.13 & 81.06 \\
    $\textbf{Compositor}^\ddag$ & ResNet-50~\cite{he2016deep} & 50M & \textbf{48.01} & \textbf{58.83} & \textbf{74.35} & \textbf{83.83} \\ \hline
    $\text{MaskFormer}^\dag$~\cite{cheng2021per} & Swin-T~\cite{liu2021swin} & 92M (46M$\times$2) & 55.42 & 67.21 & 81.37 & 89.29 \\
    $\text{MaskFormer-Dual}^{\ddag}$ & Swin-T~\cite{liu2021swin} & 51M & 54.21 & 66.42 & 81.02 & 88.71 \\
    $\textbf{Compositor}^\ddag$ & Swin-T~\cite{liu2021swin} & 51M & \textbf{55.92} & \textbf{67.63} & \textbf{83.10} & \textbf{90.42}
    \end{tabular}
    \label{tab:main_pascal}
\end{table*}

\section{Experiments}

\begin{figure*}[t!]
    \centering
    \includegraphics[width=0.95\textwidth]{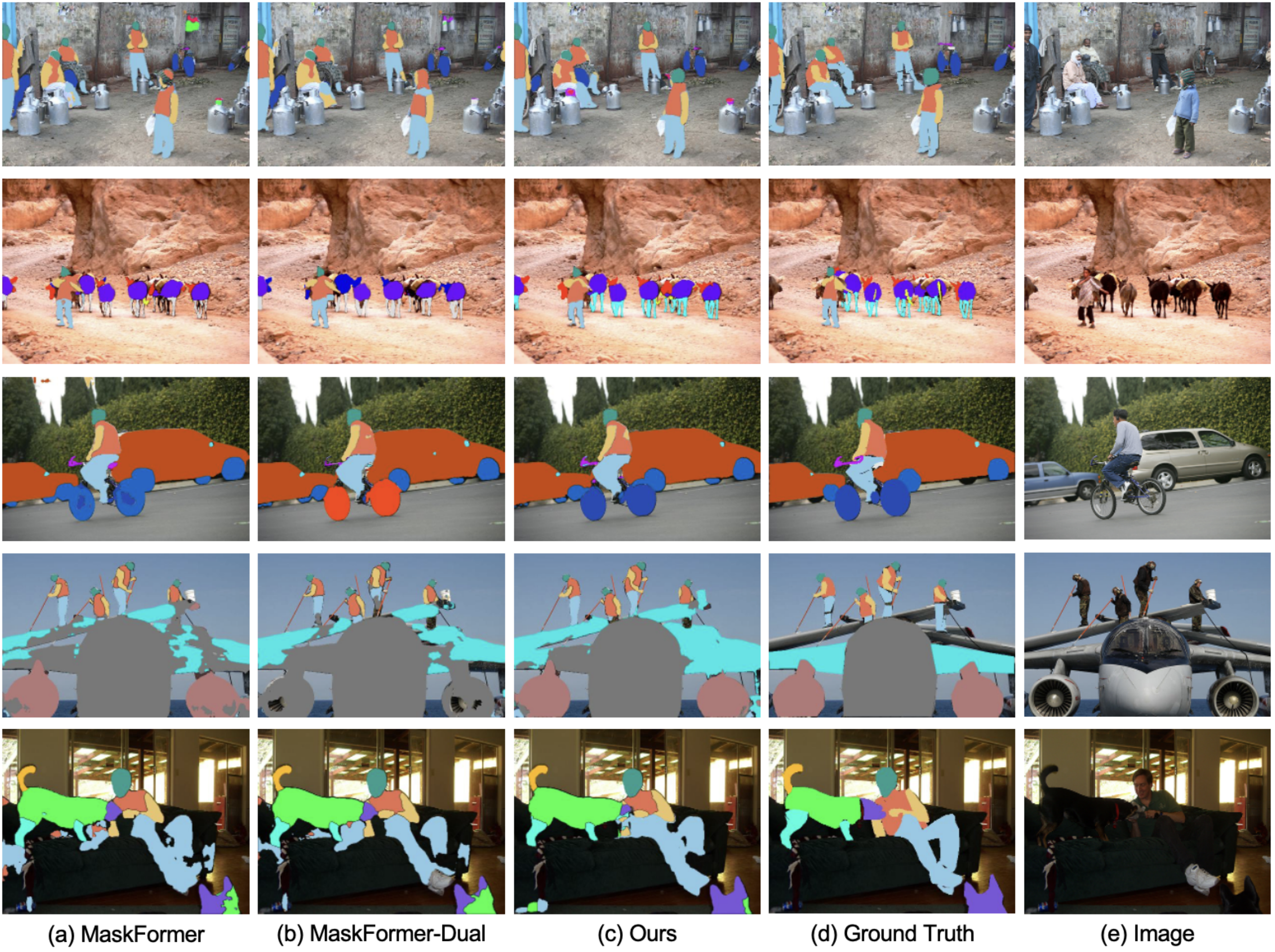}
    \caption{Qualitative comparison for different methods on Pascal-Part. Note that our Compositor produces much more accurate part segmentation results with correct labels (\eg, row 2\&3) and fewer artifacts (\eg, row 4\&5).}
    \label{fig:vis}
\end{figure*}


\textbf{Datasets.} We study Compositor on both PartImageNet \cite{he2022partimagenet} and Pascal-Part \cite{chen2014detect}. Both datasets offer large-scale and high-quality per-pixel part annotations on a wide range of objects. To be specific, PartImageNet consists of 158 classes from ImageNet \cite{deng2009imagenet} with 24,095 images. Pascal-Part is an additional annotation of VOC \cite{everingham2010pascal} which contains 10,103 images from 20 classes. For Pascal-Part, we only consider the 16 classes which have part-level annotations and ignore the rest. We manually merge the provided labels to a higher-level definition of parts (e.g. left wing \& right wing $\to$ wing) on Pascal-Part since the original parts are too fine-grained. Note that images in PartImageNet usually only contain one object while Pascal-Part scenes are more complicated with multiple objects. We follow the official train/val/test split and report performance on the val set.


\subsection{Implementation Details}

\textbf{Pixel feature extractor}. To verify the generality of Compositor across classic and more advanced feature extractor backbones, we experiment with both ResNet-50 \cite{he2016deep} and Swin Transformer \cite{liu2021swin} (tiny variants). For simplicity, a lightweight FPN \cite{kirillov2019panoptic} is adopted as the default pixel feature decoder for further enhancing the pixel embedding. Specifically, we upsample the low-resolution feature map and sum it with the corresponding upper feature map to produce a multi-scale pixel embedding at output stride $32$, $16$, and $8$ respectively. $1 \times 1$ convolution layers are used to guarantee that all per-pixel features share the same dimension. 

\textbf{\textit{Clustering} and \textit{Compositing} Steps}. The proposed \textit{Clustering} and \textit{Compositing} operations can be easily implemented with multi-head attention as illustrated in Fig. \ref{fig:framework}. Three continuous proposed \textit{Clustering} and \textit{Compositing} operations are grouped into a block to enhance the model capability. In total, three such blocks are applied to process the multi-scale pixel embeddings. The numbers of part embedding and object embedding (\ie, N, M) are set to be larger than the number of actual parts and objects to offer more proposals to the model. The specific numbers vary from different datasets and we set it to $30/5$, $50/20$ in default for PartImageNet and Pascal-Part respectively.

\textbf{Training settings.} We instantiate Compositor training setting based on MaskFormer \cite{cheng2021per}. To be specific, AdamW \cite{adamw} is adopted with an initial learning rate of 0.0002 and a weight decay of 0.05. The ImageNet-pretrained backbone has a lower learning rate with a multiplier $0.1$. We decay the learning rate by a factor of 10 at 0.9 and 0.95 fractions of the total training process. Without an additional statement, we train our models for 50k iterations on PartImageNet and 10k iterations on Pascal-Part with a batch size of 128. We adopt random cropping and large-scale jittering \cite{du2021simple, ghiasi2021simple} for data augmentations. We set $\lambda_{ce} = 5.0, \lambda_{dice} = 5.0, \lambda_{cls} = 2.5$ respectively. $\beta$ is set to be $1/2$ as default.


\subsection{Main Results}

We compare our Compositor with a few task-specialized classic CNN-based segmentation methods as well as more advanced methods with transformer architectures. Besides, we establish a dual-task version of MaskFormer by adding additional separate cross-attention modules and prediction heads on the pixel encoder, which produces both part and object segmentation results from pixels simultaneously.

Table \ref{tab:main} summarizes our experimental results on PartImageNet \cite{he2022partimagenet}. With ResNet-50 as the backbone, Compositor achieves $61.44\%$ and $71.78\%$ on part and object mIoU respectively, surpassing the task-specialized MaskFormer by around $1.6\%$. With a stronger backbone Swin-T, Compositor boosts the performance of mIoU on parts and objects to $64.64\%$ and $78.98\%$, which outperforms MaskFormer by $0.7\%$ on part and $1.1\%$ on object. However, we discover that the dual-task strategy does not bring help to these tasks, instead, it hurts the performance consistently, especially on part. We argue that this is because the dual-task pipeline ignores the connections between parts and objects, which makes these two tasks no longer beneficial to each other and increases the learning difficulty of the backbone as it needs to provide a good pixel embedding for both tasks.

Table \ref{tab:main_pascal} shows our experimental results on the Pascal-Part \cite{chen2014detect}. Compositor achieves $48.01\%$ and $74.35\%$ in terms of part and object mIoU with ResNet-50. Compositor further increases the performance to $55.92\%$ and $83.10\%$ with Swin-T, which leads MaskFormer by around $1.7\%$. Note that under multi-instances scenarios, the performance of the dual-branch model still drops by a non-trivial margin.

\begin{figure*}[t!]
    \centering
    \includegraphics[width=0.95\textwidth]{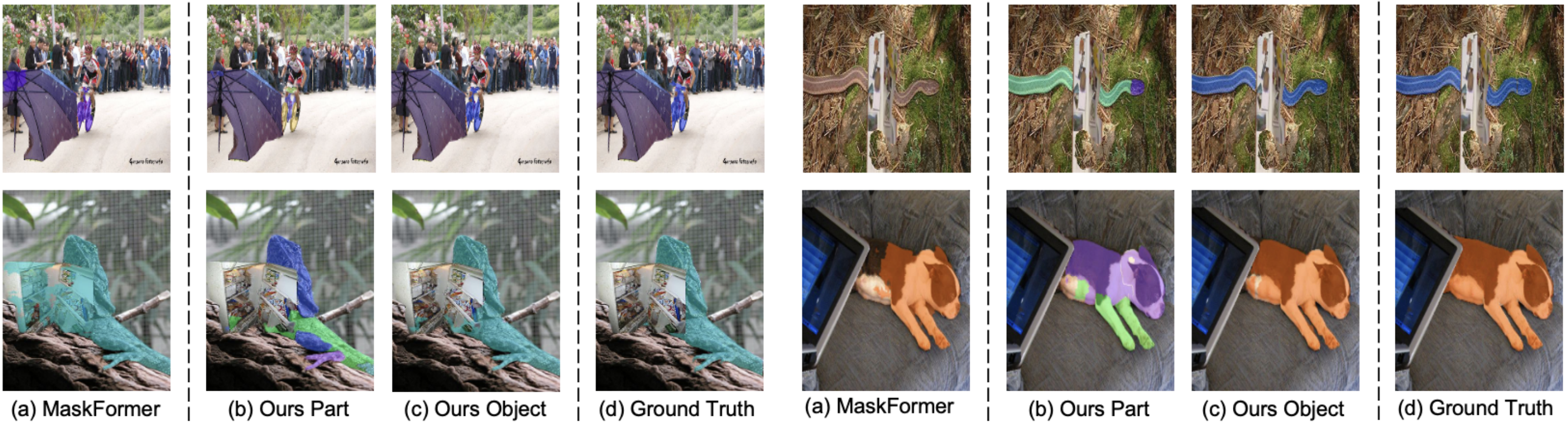}
    \caption{Qualitative comparison on Occluded-PartImageNet-v1 where we append out-of-distribution objects as occluders to randomly mask $20\% \sim 40\%$ foreground region. Compositor exhibits stronger robustness and largely avoids four kinds of common errors under occlusion: 1) wrong classification of object, 2) missing detection of object, 3) false prediction on background, 4) partial detection of object.}
    \label{fig:occ_vis}
\end{figure*}

We also conduct qualitative comparison as illustrated in Figure.~\ref{fig:vis}. We show that our Compositor is able to parse multiple different objects into corresponding parts even in complicated scenes. In general, Compositor produces more accurate part segmentation results with the correct label (\eg, row 2\&3) and fewer artifacts (\eg, row 4\&5). 

\subsection{Ablation Studies}

\textbf{Loss ratio of parts to objects}. Table \ref{tab:weight} summarizes the influence of different loss ratio $\beta$ regarding the performance on Pascal-Part. Decreasing $\beta$ at a small range can improve the performance on both tasks as better parts would bring better candidate components for objects thus also improving object segmentation. Moreover, we find that if we increase $\beta$ which stands for paying more attention to the object side, the part segmentation score drops by a large margin. 

\begin{center}
    \begin{minipage}{0.47\textwidth}
        \begin{minipage}[t]{0.48\textwidth}
            \centering
            \makeatletter\def\@captype{table}\makeatother
            \small
            \caption{Ablation study on part and object loss ratio $\beta$ with Swin-T on Pascal-Part.}
            \label{tab:weight}
            \begin{tabular}{c|c|c}
                \multirow{2}{*}{$\beta$} & Part & Object \\ \cline{2-3} 
                 & mIoU & mIoU \\ \hline
                1/4 & 55.04 & 81.97 \\
                1/3 & 55.36 & 82.65 \\
                \rowcolor{Gray}
                1/2 & \textbf{55.92} & \textbf{83.10} \\
                1 & 53.38 & 82.03 \\
                2 & 51.41 & 81.74
            \end{tabular}
        \end{minipage}
        \begin{minipage}[t]{0.48\textwidth}
            \centering
            \makeatletter\def\@captype{table}\makeatother
            \small
            \caption{Ablation study on numbers of queries (\ie, N/M) with Swin-T on Pascal-Part.}
            \label{tab:part_q}
            \begin{tabular}{c|c|c}
                \multirow{2}{*}{N/M} & Part & Object \\ \cline{2-3} 
                 & mIoU & mIoU \\ \hline
                15/40 & 54.47 & 81.74 \\
                15/50 & 54.68 & 82.03 \\
                \rowcolor{Gray}
                20/50 & \textbf{55.92} & \textbf{83.10} \\
                25/50 & 55.30 & 82.48 \\
                20/60 & 54.71 & 82.21
            \end{tabular}
      \end{minipage}
    \end{minipage}
\end{center}

\textbf{Number of queries}. Table \ref{tab:part_q} shows the influence of the number of queries. Generally speaking, the more queries we have, the more possible proposals for the part and object can be offered. Experimentally, we find that increasing the number of queries within a range boosts the performance while increasing more does not provide additional gain.



\begin{table}[]
    \small
    \centering
    \caption{Results on Occluded-PartImageNet-v1 where around $20\% \sim 40\%$ region of the object is occluded. Numbers in brackets indicate performance drop compared to that on clean images.} 
    \begin{tabular}{c|c|c}
    \multirow{2}{*}{method} & Part & Object \\ \cline{2-3} 
     & mIoU & mIoU \\ \hline
    MaskFormer \cite{cheng2021per} & 50.23 \scriptsize{(-13.73)} & 56.73 \scriptsize{(-21.19)} \\
    \textbf{Compositor (Ours)} & \textbf{54.63} \scriptsize{(-10.01)} & \textbf{63.79} \scriptsize{(-15.19)}
    \end{tabular}
\end{table}

\subsection{Occlusion}

To validate the robustness of Compositor, we propose Occluded-PartImageNet following the protocol of OccludedPASCAL3D+ \cite{wang2020robust}. Specifically, we append artificial occluders to the PartImageNet images. We create three different levels of occlusion based on the occluded ratio of the object. Compositor achieves $54.63\%$ and $63.79\%$ on part and object mIoU on Occluded-PartImageNet-v1, while the performance of MaskFormer drops more and only gets $50.23\%$ and $56.73\%$ in terms of part and object mIoU, which is around $4.4\%$ and $7.1\%$ behind Compositor. Sample visualizations are given in Fig. \ref{fig:occ_vis}, where we show how Compositor successfully segments the object through parts to avoid typical errors. We hypothesize that the robustness of Compositor comes from both the clustering idea to form parts which depends on the pixel feature similarity and the hierarchical design to composite parts into objects.

\subsection{Error Analysis}

Since our model not only predicts segmentation maps but also generates part proposals, we could further evaluate it from the perspective of part detection. Specifically, we first compute the overlap of part proposals and the groundtruth part masks. We use 0.5 as the IoU threshold to determine whether a part is detected or not (\ie, detected parts). We examine those undetected parts to see whether our model completely ignores the parts and does not make any valid prediction (\ie, missing parts) or produces part proposals at the correct region but with a wrong label (\ie, wrong parts). For those wrong parts, we check whether the correct label lies in the top-3 of the classification prediction (\ie, Top-3). As can be seen from Table \ref{tab:error_analysis}, our model successfully detects around $85\%$ and $73\%$ of parts on PartImageNet and Pascal-Part respectively. For the undetected parts, our model has valid part proposals for one fourth of them. Note that among these part proposals with wrong labels, more than $80\%$ have the correct label within top-3 prediction.

\begin{table}[]
    \small
    \centering
    \caption{Compositor part detection error pattern analysis.} 
    \begin{tabular}{cc|c|c}
    \multicolumn{2}{c|}{Error Type/Dataset} & PartImageNet & Pascal-Part \\ \hline
    \multicolumn{2}{c|}{GT Parts} & 3697 & 18392 \\ \hline
    \multicolumn{2}{c|}{Detected parts} & 3137 & 13414 \\ \hline
    \multicolumn{2}{c|}{Missing parts} & 403 & 4192 \\ \hline
    \multicolumn{1}{c|}{\multirow{2}{*}{Wrong parts}} & All & 157 & 786 \\ \cline{2-4} 
    \multicolumn{1}{c|}{} & Top-3 & 135 & 657
    \end{tabular}
    \label{tab:error_analysis}
\end{table}

%% file: 06_conclusion.tex
\section{Conclusion}
In this work, we present a bottom-up algorithm \textit{Compositor}, for jointly segmenting parts and objects by estimating part proposals from pixels which are then composed into objects. Compositor formulates semantic segmentation as an optimization problem which involves learning feature embeddings so that parts and objects can be represented by centroids in this embedding space. They are estimated, respectively, by
\textit{Clustering} and \textit{Compositing} lower semantic level embeddings. Compositor achieves state-of-art-performance on both part and object segmentation on PartImageNet and Pascal-Part and shows stronger robustness against occlusion on our proposed Occluded-PartImageNet. Error analysis is provided to better understand our model.

\noindent \textbf{Acknowledgements.} The authors gratefully acknowledge
supports from ONR with award N00014-21-1-2812 and IAA with award 80052272.